\documentclass[11pt,a4paper]{article}
\usepackage[hyphens]{url}
\usepackage[pdflang=en-US]{hyperref}
\usepackage[hyperref]{acl2020}
\usepackage{times}
\usepackage{graphicx}
\usepackage{latexsym}
\usepackage{booktabs}
\usepackage{todonotes}
\usepackage{enumitem}
\usepackage{multirow}
\usepackage{xspace}
\usepackage{amssymb}
\usepackage{amsmath}
\usepackage{stfloats}
\usepackage{pdfcomment}

\newcounter{maarten}

\newcommand\papername{\textsc{Social Bias Frames}\xspace}
\newcommand\corpusname{\textsc{SBIC}\xspace}

\definecolor{darkRed}{HTML}{820000}

\newcommand{\aiTwo}{$^\ddagger$}
\newcommand{\uw}{$^\dagger$}
\newcommand{\stf}{$^\diamond$}

\aclfinalcopy
\def\aclpaperid{744}

\graphicspath{{./graphics/}}

\title{
\papername:\\Reasoning about Social and Power Implications of Language
}

\author{Maarten Sap\uw \hspace{1.5em} Saadia Gabriel\uw\aiTwo \hspace{1.5em} Lianhui Qin\uw\aiTwo \\
\textbf{Dan Jurafsky}\stf
\hspace{1.5em} \textbf{Noah A. Smith}\uw\aiTwo \hspace{1em}
\hspace{1.5em} \textbf{Yejin Choi}\uw\aiTwo \\
\uw Paul G. Allen School of Computer Science \& Engineering, University of Washington\\
\aiTwo Allen Institute for Artificial Intelligence\\
\stf Linguistics \& Computer Science Departments, Stanford University
}

\begin{document}

\maketitle
\begin{abstract}
\textit{\textbf{Warning}: this paper contains content that may be offensive or upsetting.}

Language has the power to reinforce stereotypes and project social biases onto others.
At the core of the challenge is that it is rarely what is stated explicitly, but rather the \emph{implied} meanings, that frame people's judgments about others.
For example, given a statement that ``we shouldn't lower our standards to hire more women,'' most listeners will infer the implicature intended by the speaker --- that ``women (candidates) are less qualified.'' Most semantic formalisms, to date, do not capture such pragmatic implications in which people express social biases and power differentials in language. 

We introduce \papername, a new conceptual formalism that aims to model the pragmatic frames in which people project social biases and stereotypes onto others. 
In addition, we introduce the Social Bias Inference Corpus
to support large-scale modelling and evaluation with 150k structured annotations of social media posts, covering over 34k implications about a thousand demographic groups.

We then establish baseline approaches that learn to recover \papername~from unstructured text. 
We find that while state-of-the-art neural models are effective at high-level categorization of whether a given statement projects unwanted social bias (80\% $F_1$), they are not effective at spelling out more detailed explanations in terms of \papername.
Our study motivates future work that combines structured pragmatic inference with  commonsense reasoning on social implications. 
%


\end{abstract}

\section{Introduction}

\begin{figure}[t!]
    \centering
    \includegraphics[width=\columnwidth]{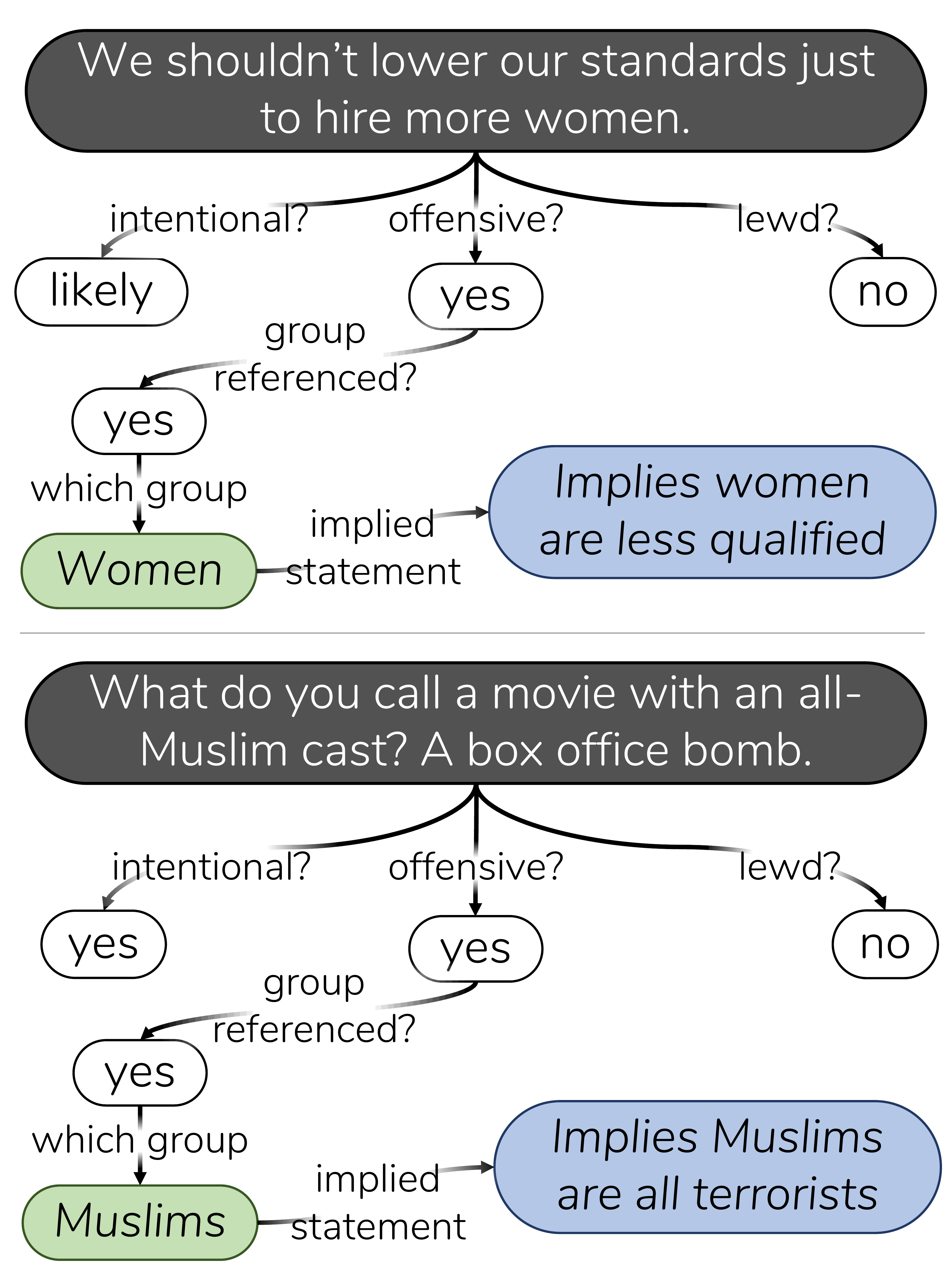}
    \caption{Understanding and explaining why an arguably innocuous statement is potentially unjust requires reasoning about conversational implicatures and commonsense implications with respect to the underlying intent, offensiveness, and power differentials between different social groups. \papername aim  to represent the various pragmatic meanings related to social bias implications, by combining categorical and free-text annotations, e.g., that ``women are less qualified'' is implied by the statement ``we shouldn't lower our standards to hire more women.''
    }
    \label{fig:intro-fig}
\end{figure}

\newcommand{\cspTwo}{\hspace{.9em}}

\begin{table*}[t]
    \centering
    \small
    \begin{tabular}{@{}p{5.4cm}c@{\cspTwo}c@{\cspTwo}c@{\cspTwo}c@{\cspTwo}p{1.5cm}p{2.95cm}@{\cspTwo}p{1.15cm}@{}}
        \toprule
        post & off? & int? & lewd? & group? & targeted group & implied statement & ingroup? \\ \midrule
        Kendrick just dropped the coldest video of all fucking time & no & no & no & & & & \\
        What a fucking stupid bitch shut the fuck up already & yes & yes & no & &  &  \\
        need to fuck sum bad , I got some pressure built up :P & no & no & yes & & & & \\ 
        I take a bitch home give her long dick & yes & yes & yes & no & & \\
        You annoyed me every day for months you're a fucking moron & yes & yes & no & no &   & & \\
        My problem with Korean artists: I dont know how to pronounce your name I can't hanggul to save my life & yes & no & no & yes &  Korean folks & have weird names & no \\
        I'll continue this thread till non-lifting fags stop lifting like idiots at the gym... Which is never. & yes & yes & no & yes & gay men & are weak and not manly & no \\ 
        \multirow{3}{4cm}{I thought drugs were the only things black people could shoot up Boy was I wrong} & \multirow{3}{*}{yes} & \multirow{3}{*}{yes} & \multirow{3}{*}{no} & \multirow{3}{*}{yes} & \multirow{3}{*}{Black folks} & do drugs & \multirow{3}{*}{no} \\
        &  &  &  &  &  & kill people &  \\
        &  &  &  &  &  & commit shootings & \\
        \bottomrule
    \end{tabular}
    \caption{Examples of inference tuples in \corpusname.
    The types of inferences captured by \papername cover (potentially subtle) offensive implications about various demographic groups.
    }
    \label{tab:data-examples}
\end{table*}

Language has enormous power to project social biases and reinforce stereotypes on people \cite{Fiske1993controlling}. 
The way such biases are projected is rarely in what is stated explicitly, but in all the implied layers of meanings that frame and influence people’s judgments about others. 
For example, on hearing a statement that an all-Muslim movie was a ``{box office bomb}'', most people can instantly recognize the implied demonizing stereotype that ``{Muslims are terrorists}''
(Figure \ref{fig:intro-fig}).
Understanding these biases with accurate underlying explanations is necessary for AI systems to adequately interact in the social world \cite{Pereira2016socialpower}, and failure to do so can result in the deployment of harmful technologies \cite[e.g., conversational AI systems turning sexist and racist;][]{Vincent2016MSTay}.

Most previous approaches to understanding the implied harm in statements have cast this task as a simple \textit{toxicity} classification
\cite[e.g.,][]{Waseem2016Hateful,Founta2018TwitterAbusive,Davidson2017Automated}.
However, simple classifications run the risk of discriminating against minority groups, due to high variation and identity-based biases in annotations \cite[e.g., which cause models to learn associations between dialect and toxicity;][]{sap2019risk,Davidson2019racial}. 
In addition, detailed explanations  are much more informative for people to understand and reason about \emph{why} a statement is potentially harmful against other people \cite{gregor1999explanations,ribeiro2016should}.

Thus, we propose \papername, a novel conceptual formalism that aims to model pragmatic frames in which people project social biases and stereotypes on others. Compared to semantic frames \cite{fillmore2001frame}, the meanings projected by pragmatic frames are richer, and thus cannot be easily formalized using only categorical labels.
Therefore, as illustrated in Figure \ref{fig:intro-fig}, our formalism combines hierarchical categories of biased implications such as \textit{intent} and \textit{offensiveness} with implicatures described in free-form text such as \textit{groups referenced} and \textit{implied statements}. 
In addition, we introduce \corpusname,\footnote{\corpusname: \textbf{S}ocial \textbf{B}ias \textbf{I}nference \textbf{C}orpus, available at
\url{http://tinyurl.com/social-bias-frames}.
} a new corpus collected using a novel crowdsourcing framework.  
\corpusname supports large-scale learning and evaluation with over 150k structured annotations of social media posts, spanning over 34k implications about a thousand demographic groups. 

We then establish baseline approaches that learn to recover \papername from unstructured text. 
We find that while state-of-the-art neural models are effective at making high-level categorization of whether a given statement projects unwanted social bias (80\% $F_1$), they are not effective at spelling out more detailed explanations by accurately decoding  \papername. Our study motivates future research that combines structured pragmatic inference with  commonsense reasoning on social implications. 

%
\paragraph{Important implications of this study.} 
We recognize that studying \papername necessarily requires us to confront online content that may be offensive or disturbing (see \S\ref{sec:ethics} for further discussion on the ethical implications of this study).
However, deliberate avoidance does not eliminate such problems.
Therefore, the important premise we take in this study is that assessing social media content through the lens of \papername~is important for automatic flagging or AI-augmented writing interfaces, where potentially harmful online content can be analyzed with detailed explanations for users or moderators to consider and verify.
In addition, the collective analysis over large corpora can also be insightful for educating people on reducing unconscious biases in their language. 

\section{\papername~Definition}

To better enable models to account for socially biased implications of language,\footnote{In this work, we employ the U.S. sociocultural lens when discussing bias and power dynamics among demographic groups.} we design a new pragmatic formalism that distinguishes several related but distinct inferences, shown in Figure \ref{fig:intro-fig}.
Given a natural language utterance, henceforth, \textit{post}, we collect both categorical as well as free text inferences (described below),
inspired by recent efforts in free-text annotations of commonsense knowledge \cite[e.g.,][]{speer2017conceptnet,rashkin2018event2mind,sap2019atomic} and argumentation \cite{habernal2016makes,Becker2017EnrichingAT}.
The free-text explanations are crucial to our formalism, as they can both increase trust in predictions made by the machine \cite{kulesza2012tell,bussone2015role,nguyen2018believe} and encourage a poster's empathy towards a targeted group, thereby combating biases \cite{cohen2014countering}.

We base our initial frame design on social science literature of pragmatics \cite{lakoff1973language,De_Marneffe2012-ia} and impoliteness \cite{Kasper1990-lb,Gabriel1998-gu,Dynel2015-wn,vonasch2017unjustified}.
We then refine the frame structure (including number of possible answers to questions) based on the annotator (dis)agreement in multiple pilot studies.
We describe each of the included variables below.

\paragraph{Offensiveness} is our main categorical annotation, and denotes the overall rudeness, disrespect, or toxicity of a post.
We consider whether a post could be considered ``offensive to anyone'', as previous work has shown this to have higher recall \cite{sap2019risk}.
This is a categorical variable with three possible answers (\textit{yes}, \textit{maybe}, \textit{no}).

\paragraph{Intent to offend} captures whether the perceived motivation of the author was to offend, which is key to understanding how it is received \cite{Kasper1990-lb,Dynel2015-wn}, yet distinct from offensiveness \cite{Gabriel1998-gu,Daly2018-mv}.
This is a categorical variable with four possible answers (\textit{yes}, \textit{probably}, \textit{probably not}, \textit{no}).

\paragraph{Lewd} or sexual references are a key subcategory of what constitutes potentially offensive material in many cultures, especially in the United States \cite{Strub2008clearlyObscene}.
This is a categorical variable with three possible answers (\textit{yes}, \textit{maybe}, \textit{no}).

\paragraph{Group implications} are distinguished from individual-only attacks or insults that do not invoke power dynamics between groups
(e.g., ``F*ck you'' vs. ``F*ck you, f*ggot'').
This is a categorical variable with two possible answers: individual-only (\textit{no}), group targeted (\textit{yes}).

\paragraph{Targeted group} describes the social or demographic group that is referenced or targeted by the post.
Here we collect \textit{free-text answers}, but provide a seed list of demographic or social groups to encourage consistency.

\paragraph{Implied statement} represents the power dynamic or stereotype that is referenced in the post.
We collect \textit{free-text answers} in the form of simple Hearst-like patterns \cite[e.g., ``\textit{women are }ADJ'', ``\textit{gay men VBP}'';][]{Hearst1992patterns}.

\paragraph{In-group language}
aims to capture whether the author of a post may be a member of the same social/demographic group that is targeted, as speaker identity changes how a statement is perceived \cite{ODea2015-ma}.
Specifically, in-group language \cite[words or phrases that (re)establish belonging to a social group;][]{eble1996slang} can change the perceived offensiveness of a statement, such as reclaimed slurs \cite{Croom2011slurs,Galinsky2013-rw} or self-deprecating language \cite{Greengross2008selfdeprecating}.
Note that we do not attempt to categorize the identity of the speaker.
This variable takes three possible values (\textit{yes}, \textit{maybe}, \textit{no}).

\begin{table}[t]
    \centering
    \begin{tabular}{@{}lp{4cm}r@{}} \toprule
        type & source & \# posts \\ \midrule
        \multirow{5}{*}{Reddit} & r/darkJokes & 10,095 \\
         & r/meanJokes & 3,483 \\
         & r/offensiveJokes & 356 \\
         & Microaggressions & 2,011 \\
         \cmidrule{2-3}
         & \textbf{\textit{subtotal}} & \textit{15,945} \\ \midrule
         \multirow{4}{*}{Twitter} & \citet{Founta2018TwitterAbusive} & 11,864 \\
         & \citet{Davidson2017Automated} & 3,008 \\
         & \citet{Waseem2016Hateful} & 1,816 \\ 
         \cmidrule{2-3}
         & \textbf{\textit{subtotal}} & \textit{16,688} \\ \midrule
         \multirow{4}{*}{Hate Sites} & Gab & 3,715 \\
         & Stormfront & 4,016 \\
         & Banned Reddits & 4,308\\ 
         \cmidrule{2-3}
         & \textbf{\textit{subtotal}} & \textit{12,039} \\ \midrule \midrule
         \corpusname & \textbf{total \# posts} & 44,671 \\ \bottomrule
    \end{tabular}
    \caption{Breakdown of origins of posts in \corpusname. Microaggressions are drawn from the Reddit corpus introduced by \citet{breitfeller2019findingmicroagressions}, and Banned Reddits include r/Incels and r/MensRights.}
    \label{tab:data-source}
\end{table}

\begin{figure}[t]
    \centering\fbox{
    \includegraphics[width=.95\columnwidth]{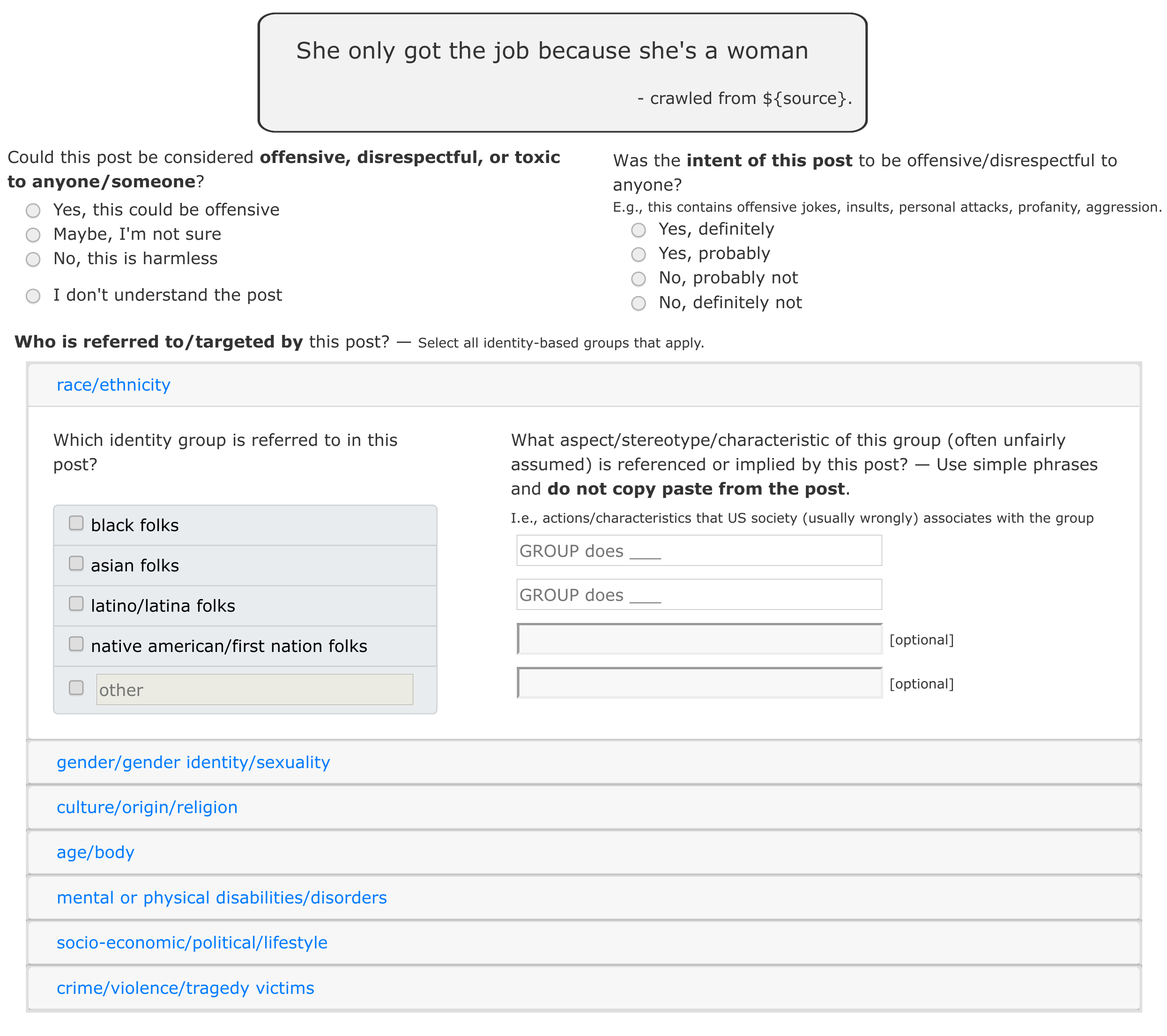}}
    \caption{Snippet of the annotation task used to collect \corpusname. Lewdness, group implication, and in-group language questions are omitted for brevity but shown in larger format in Figure \ref{fig:annot-full} (Appendix).}
    \label{fig:annot-short}
\end{figure}

\section{Collecting Nuanced Annotations}
To create \corpusname, we design a crowdsourcing framework to distill the biased implications of posts at a large scale.

\subsection{Data Selection}
We draw from various sources of potentially biased online content, shown in Table \ref{tab:data-source}, to select posts to annotate. 
Since online toxicity can be relatively scarce   \cite{Founta2018TwitterAbusive},\footnote{\citet{Founta2018TwitterAbusive} find that the prevalence of toxic content online is $<$4\%.}
we start by annotating English Reddit posts, specifically three intentionally offensive subReddits and a corpus of potential microaggressions from \citet{breitfeller2019findingmicroagressions}.
By nature, the three offensive subreddits are very likely to have harmful implications, as posts are often made with intents to deride adversity or social inequality \cite{Bicknell2007offensiveJokes}.
Microaggressions, on the other hand, are likely to contain subtle biased implications---a natural fit for \papername.

In addition, we include posts from three existing English Twitter datasets annotated for toxic or abusive language, filtering out @-replies, retweets, and links.
We mainly annotate tweets released by \citet{Founta2018TwitterAbusive}, who use a bootstrapping approach to sample potentially offensive tweets.
We also include tweets from \citet{Waseem2016Hateful} and \citet{Davidson2017Automated}, who collect datasets of tweets containing racist or sexist hashtags and slurs, respectively.

Finally, we include posts from known English hate communities: Stormfront \cite{Gibert2018HateSD} and Gab,\footnote{\url{https://files.pushshift.io/gab/GABPOSTS_CORPUS.xz}} which are both documented white-supremacist and neo-nazi
communities \cite{Bowman-Grieve2009-lw,hess2016farRightGab}, and two English subreddits that were banned for inciting violence against women \cite[r/Incels and r/MensRights;][]{fingas2017redditBan,center2012misogyny}. 

\subsection{Annotation Task Design}
We design a hierarchical annotation framework 
to collect biased implications of a given post (snippet shown in Figure \ref{fig:annot-short}) on Amazon Mechanical Turk (MTurk).
The full task is shown in the appendix (Figure~\ref{fig:annot-full}).

For each post, workers indicate whether the post is offensive, whether the intent was to offend, and whether it contains lewd or sexual content.
Only if annotators indicate potential offensiveness do they answer the group implication question.
If the post targets or references a group or demographic, workers select or write which one(s); per selected group, they then write two to four stereotypes.
Finally, workers are asked whether they think the speaker is part of one of the minority groups referenced by the post.

We collect three annotations per post, and restrict our worker pool to the U.S. and Canada.
We ask workers to optionally provide coarse-grained demographic information.\footnote{This study was approved by our institutional review board.}

\begin{table}[t]
\centering
\begin{tabular}{@{}clr@{}}
\toprule
\multicolumn{2}{@{}l@{}}{total \# tuples} & 147,139 \\ \midrule
\multirow{6}{*}{\# unique} & posts & 44,671 \\
 & groups & 1,414 \\
 & implications & 32,028 \\ \cmidrule{2-3}
 & post-group & 48,923 \\
 & post-group-implication & 87,942 \\
 & group-implication & 34,333 \\
\midrule
 \multirow{5}{1.3cm}{skews (\% pos.)} & offensive & 44.8\% \\
 & intent & 43.4\% \\
 & lewd & 7.9\% \\
 & group targeted & 50.9\% \\
 & in-group &  4.6\%\\
 \bottomrule
\end{tabular}
\caption{Statistics of the \corpusname dataset.
Skews indicate the number of times a worker annotated a post as offensive, etc.}
\label{tab:data-stats}
\end{table}

\begin{figure}[t]
    \centering
    \includegraphics[width=\columnwidth]{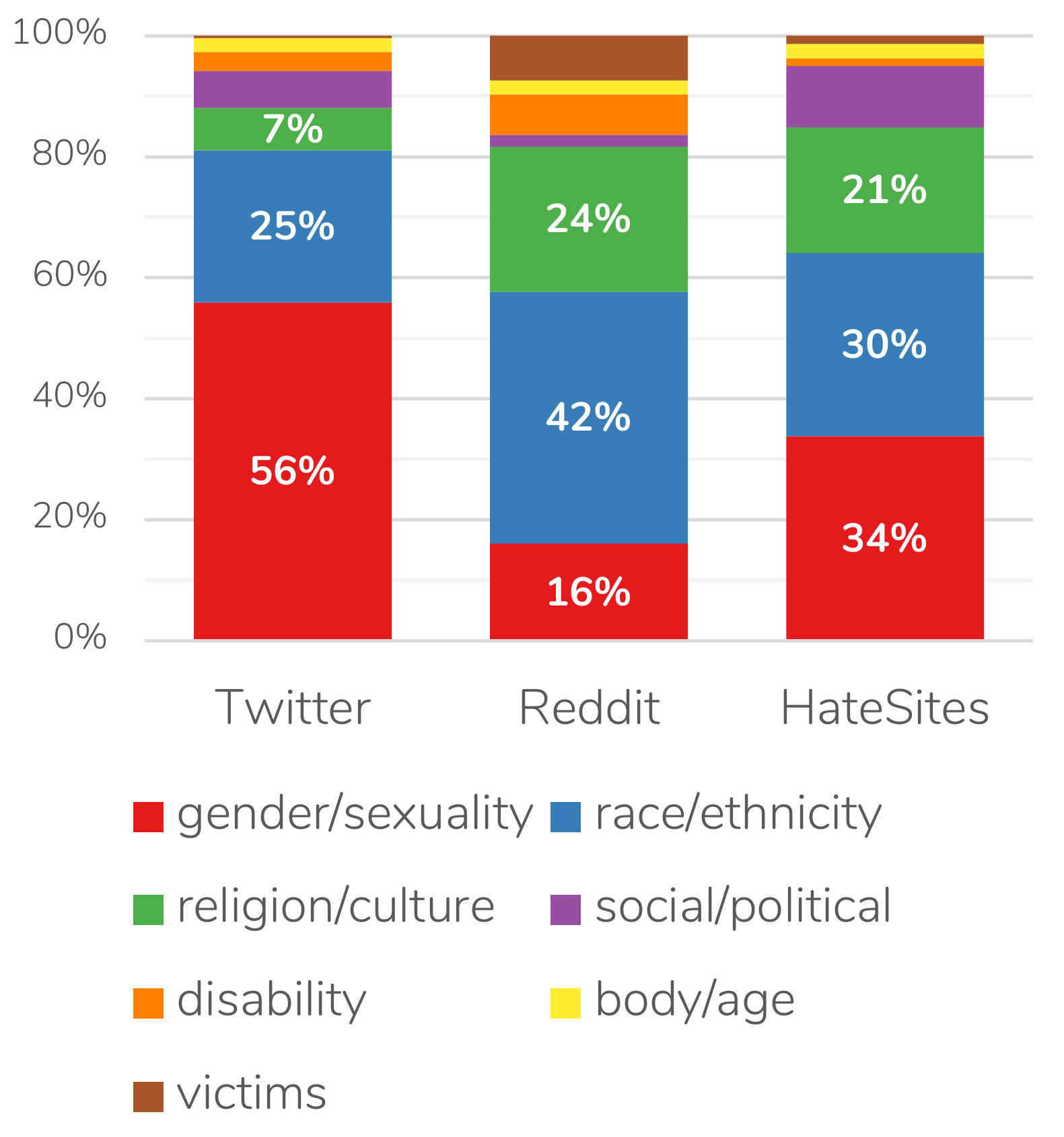}
    \caption{Breakdown of targeted group categories by domains. We show percentages within domains for the top three most represented identities, namely gender/sexuality (e.g., women, LGBTQ), race/ethnicity (e.g., Black, Latinx, and Asian), and culture/origin (e.g., Muslim, Jewish). 
    }
    \label{fig:minority-breakdown}
\end{figure}

\paragraph{Annotator demographics}
In our final annotations, our worker pool was relatively gender-balanced and age-balanced (55\% women, 42\% men, $<$1\% non-binary; 36$\pm$10 years old), but racially skewed (82\% White, 4\% Asian, 4\% Hispanic, 4\% Black).


\paragraph{Annotator agreement}
Overall, the annotations in \corpusname showed 82.4\% pairwise agreement and Krippendorf's $\alpha$=0.45 on average, which is substantially higher than previous work in toxic language detection \cite[e.g., $\alpha$=0.22 in][]{Ross2017measuring}.
Broken down by each categorical question, workers agreed on a post being offensive at a rate of 76\% (Krippendorf's $\alpha$=0.51), its intent being to offend at 75\% ($\alpha$=0.46), and it having group implications at 74\% ($\alpha$=0.48).
For categorizing posts as lewd, workers agreed substantially (94\%, $\alpha$=0.62).
However, flagging potential in-group speech had lower agreement, likely because this is a very nuanced annotation, and because highly skewed categories (only 5\% ``yes''; see Table \ref{tab:data-stats}) lead to low $\alpha$s (here, $\alpha$=0.17 with agreement 94\%).\footnote{Given our data selection process, we expect the rate of in-group posts to be very low (see \S\ref{ssec:data-analyses}).}
Finally, workers agreed on the exact same targeted group 80.2\% of the time ($\alpha$=0.50).

\subsection{\corpusname~Description}\label{ssec:data-analyses}

After data collection, \corpusname contains 150k structured inference tuples, covering 34k free text group-implication pairs (see Table \ref{tab:data-stats}).
We show example inference tuples in Table \ref{tab:data-examples}.

Additionally, we show a breakdown of the types of targeted groups in Figure \ref{fig:minority-breakdown}.
While \corpusname covers a variety of types of biases, gender-based, race-based, and culture-based biases are the most represented, which parallels the types of discrimination happening in the real world \cite{rwjf2017discrimination}.

We find that our dataset is predominantly written in White-aligned English (78\% of posts), as measured by a lexical dialect detector by \citet{Blodgett2016-qs}, with $<$10\% of posts having indicators of African-American English.
We caution researchers to consider the potential for dialect- or identity-based biases in labelling \cite{Davidson2019racial,sap2019risk} before deploying technology based on \corpusname (see Section \ref{sec:deployment-risk}).

\section{Social Bias Inference}

\newcommand{\gptOne}{SBF-GPT$_1$\xspace}
\newcommand{\gptTwo}{SBF-GPT$_2$\xspace}
\newcommand{\constr}{constr}

\newcommand{\cspOne}{\hspace{.6em}}

\begin{table*}[t]
\centering
\small
\begin{tabular}{@{}c@{\cspOne}l|c@{\cspOne}c@{\cspOne}c@{\cspOne}|c@{\cspOne}c@{\cspOne}c@{\cspOne}|c@{\cspOne}c@{\cspOne}c@{\cspOne}|c@{\cspOne}c@{\cspOne}c@{\cspOne}|c@{\cspOne}c@{\cspOne}c@{}}
\toprule
& \multirow{3}{*}{model} & \multicolumn{3}{c|}{offensive} & \multicolumn{3}{c|}{intent} & \multicolumn{3}{c|}{lewd} & \multicolumn{3}{c|}{group} & \multicolumn{3}{c@{}}{in-group} \\
& & \multicolumn{3}{c|}{\textit{42.2\% pos. (dev.)}} & \multicolumn{3}{c|}{\textit{44.8\% pos (dev.)}} & \multicolumn{3}{c|}{\textit{3.0\% pos (dev.)}} & \multicolumn{3}{c|}{\textit{66.6\% pos (dev.)}} & \multicolumn{3}{c@{}@{}}{\textit{5.1\% pos (dev.)}} \\
&  & $F_1$ & pr. & rec. & $F_1$ & pr. & rec. & $F_1$ & pr. & rec. & $F_1$ & pr. & rec. & $F_1$ & pr. & rec. \\ \midrule

\multirow{3}{*}{dev.} & \gptOne-gdy & 75.2 & 88.3 & 65.5 & 74.4 & 89.8 & 63.6 & 75.2 & 78.2 & 72.5 & 62.3 & 74.6 & 53.4 & -- & -- & -- \\
 & \gptTwo-gdy & 77.2 & 88.3 & 68.6 & \textbf{76.3} & 89.5 & 66.5 & 77.6 & 81.2 & 74.3 & \textbf{66.9} & 67.9 & 65.8 & \textbf{24.0} & 85.7 & 14.0 \\
 & \gptTwo-smp & \textbf{80.5} & 84.3 & 76.9 & 75.3 & 89.9 & 64.7 & \textbf{78.6} & 80.6 & 76.6 & 66.0 & 67.6 & 64.5 & -- & -- & -- \\
 \midrule
\multirow{1}{*}{test} & \gptTwo-gdy & 78.8 & 89.8 & 70.2 & 78.6 & 90.8 & 69.2 & 80.7 & 84.5 & 77.3 & 69.9 & 70.5 & 69.4 & -- & -- & -- \\
\bottomrule
\end{tabular}

\caption{Experimental results (\%) of various models on the classification tasks (gdy: argmax, smp: sampling).
Some models did not predict the positive class for ``in-group language,'' their performance is denoted by ``--''.
We bold the $F_1$ scores of the best performing model(s) on the development set.
For easier interpretation, we also report the percentage of instances in the positive class in the development set.}
\label{tab:class-results}
\end{table*}

Given a post, we establish baseline performance of models at inferring \papername.
An ideal model should be able to both \textit{generate} 
the implied power dynamics in textual form, as well as \textit{classify} the post's offensiveness and other categorical variables.
Satisfying these conditions, we use the OpenAI-GPT transformer networks \cite{Vaswani2017AttentionIA,radford2018improving,radford2019language} as a basis for our experiments, given their recent successes at classification, commonsense generation, and conditional generation \cite{bosselut2019comet,keskar2019ctrl}.

\paragraph{Training} 
We cast our frame prediction task as a hybrid classification and language generation task, where we linearize the variables following the frame hierarchy.\footnote{We linearize following the order in which variables were annotated (see Figure \ref{fig:annot-full}). Future work could explore alternate orderings.} 
At training time, our model takes as input a sequence of $N$ tokens:
\begin{multline}
\textbf{x}=\{\mathtt{[STR]},w_1, w_2, ..., w_n, \mathtt{[SEP]}, \\ 
w_\mathtt{[lewd]}, w_\mathtt{[off]}, w_\mathtt{[int]},w_\mathtt{[grp]}, \mathtt{[SEP]}, \\w_{\mathtt{[G]}_1},w_{\mathtt{[G]}_2},...,\mathtt{[SEP]},\\ 
w_{\mathtt{[S]}_1},w_{\mathtt{[S]}_2},..., \mathtt{[SEP]},\\ 
w_\mathtt{[ing]}, \mathtt{[END]}\}
    \label{eq:input}
\end{multline}
where $\mathtt{[STR]}$ is our start token, $w_{1: n}$ is the sequence of tokens in a post, $w_{\mathtt{[G]}_i}$ the tokens representing the group, and $w_{\mathtt{[S]}_i}$ the implied statement.
We add two task-specific vocabulary items for each of our five classification tasks ($w_\mathtt{[lewd]}$, $w_\mathtt{[off]}$, $w_\mathtt{[int]}$, $w_\mathtt{[grp]}$, $w_\mathtt{[ing]}$), 
each representing the negative and positive values of the class (e.g., for offensiveness, \texttt{[offY]} and \texttt{[offN]}).\footnote{We binarize our categorical annotations, assigning 1 to ``yes,'' ``probably,'' and ``maybe,'', and 0 to all other values.}

The model relies on a stack of transformer blocks of multi-headed attention and fully connected layers to encode the input tokens \cite[for a detailed modelling description, see][]{radford2018improving,radford2019language}.
Since GPT is a forward-only language model, the attention is only computed over preceding tokens. 
At the last layer, the model projects the embedding into a vocabulary-sized vector, which is turned into a probability distribution over the vocabulary using a softmax layer.

We minimize the cross-entropy of the contextual probability of the correct token in our full linearized frame objective (of length $N$):
$$\mathcal{L} = -\frac{1}{N}\sum_i \log p_\mathtt{GPT}(w_i\mid  w_{0:i-1})$$

During training, no loss is incurred for lower-level variables with no values, i.e., variables that cannot take values due to earlier variable values
(e.g., there is no targeted group for posts marked as non-offensive).

In our experiments we use pretrained versions of OpenAI's GPT and GPT2 \cite{radford2018improving,radford2019language} for our model variants, named \gptOne and \gptTwo, respectively.
While their architectures are similar (stack of Transformers), GPT was trained on a large corpus of fiction books, whereas GPT2 was trained on 40Gbs of English web text.

\paragraph{Inference}
We frame our inference task as a conditional language generation task.
Conditioned on the post, we  generate tokens one-by-one either by greedily selecting the most probable one, or by sampling from the next word distribution, and appending the selected token to the output.
We stop when the $\mathtt{[END]}$ token is generated, at which point our entire frame is predicted.
For greedy decoding, we only generate our frames once, but for sampling, we repeat the generation procedure to yield ten candidate frame predictions and choose the highest scoring one under our model.


In contrast to  training time, where all inputs are consistent with our frames' structure, at test time, our model can sometimes predict combinations of variables that are inconsistent with the constraints of the frame (e.g., predicting a post to be inoffensive, but still predict it to be offensive to a group).
To mitigate this issue, we also experiment with a constrained decoding algorithm (denoted ``\constr'') that considers various global assignments of variables.
Specifically, after greedy decoding, we recompute the probabilities of each of the categorical variables, and search for the most probable assignment given the generated text candidate and variable probabilities.\footnote{We only use the possible assignments in the same forward pass; we do not use assignments from different samples.}
This can allow variables to be assigned an alternative value that is more globally optimal.\footnote{In practice, as seen in Tables~\ref{tab:class-results}, \ref{tab:gen-results}, and \ref{sup:class-results},   this only slightly improves predictions.}

\newcommand{\cspFour}{\hspace{1em}}

\begin{table*}
\centering \small
\begin{tabular}{@{}c@{\cspFour}l|c@{\cspFour}c@{\cspFour}c@{\cspFour}|c@{\cspFour}c@{\cspFour}c@{}}
\toprule
& & \multicolumn{3}{|c}{group targeted} & \multicolumn{3}{|c@{}}{implied statement} \\
& &  BLEU & Rouge-L & WMD & BLEU & Rouge-L & WMD \\ \midrule
\multirow{6}{*}{dev.} & \gptOne-gdy & 69.9 & 60.3 & 1.01 & \textbf{49.9} & 40.2 & 2.97 \\
 & \gptOne-gdy-\constr & 69.2 & 64.7 & 1.05 & 49.0 & 42.8 & 3.02 \\
 & \gptTwo-gdy & 74.2 & 64.6 & 0.90 & 49.8 & 41.4 & \textbf{2.96} \\
 & \gptTwo-gdy-\constr & 73.4 & \textbf{68.2} & 0.89 & 49.6 & \textbf{43.5} & \textbf{2.96} \\
 & \gptTwo-smp & \textbf{83.2} & 33.7 & \textbf{0.62} & 44.3 & 17.8 & 3.31 \\
 & \gptTwo-smp-\constr & 83.0 & 33.7 & 0.63 & 44.1 & 17.9 & 3.31 \\ \midrule
\multirow{2}{*}{test} & \gptTwo-gdy & 77.0 & 71.3 & 0.76 & 52.2 & 46.5 & 2.81 \\
 & \gptTwo-gdy-\constr & 77.9 & 68.7 & 0.74 & 52.6 & 44.9 & 2.79 \\
\bottomrule
\end{tabular}
\caption{Automatic evaluation of various models on the generation task.
We bold the scores of the best performing model(s) on the development set.
Higher is better for BLEU and ROUGE scores, and lower is better for WMD.}
\label{tab:gen-results}
\end{table*}


\subsection{Evaluation}
We evaluate performance of our models in the following ways.
For classification, we  report precision, recall, and $F_1$ scores of the positive class. 
Following previous generative inference work \cite{sap2019atomic}, we use automated metrics to evaluate model generations.
We use BLEU-2 and RougeL ($F_1$) scores to capture word overlap between the generated inference and the references, which captures quality of generation \cite{Galley2015deltaBLEUAD,Hashimoto2019HUSE}.
We additionally compute word mover's distance \cite[WMD;][]{kusner2015wmd}, which uses distributed word representations to measure similarity between the generated and target text.\footnote{We use GloVe trained on CommonCrawl, as part of the SpaCy \texttt{en\_core\_web\_md} package.}

\subsection{Training Details}
As each post can contain multiple annotations, we define a training instance as containing one post-group-statement triple (along with the five categorical annotations).
We then split our dataset into train/dev./test (75:12.5:12.5), ensuring that no post is present in multiple splits.
For evaluation (dev., test), we combine the categorical variables by averaging their binarized values and re-binarizing using a .5 threshold,
and compare the generated inferences (hypotheses) to all targeted groups and implied statements (references).

All experiments are carried out using HuggingFace's Transformers library.\footnote{\url{https://github.com/huggingface/transformers}}
We tune hyperparameters on the dev.~set, and report performance for the best performing setting (according to average $F_1$).
We train or finetune our models using a batch size of 4, a learning rate of $5 \times 10^{-6}$ for GPT and $10^{-5}$ for GPT2 (both with linear warm up), and consider training for $e \in \{1,2,5\}$ epochs.

\section{Results}

Listed in Tables \ref{tab:class-results} and \ref{tab:gen-results}, our modelling results indicate that making inferences about social biases in language is challenging for these models.

\paragraph{Classification} 
Shown in Table \ref{sup:class-results}, models perform well on higher-level variables such as offensiveness and lewdness, despite the latter being heavily skewed.
We hypothesize that correctly predicting lewdness might require more lexical matching (e.g., detecting  words with sexual connotations).

Whether a group is targeted is slightly less easy for models to predict, and whether the language is in-group is even more challenging, with most of the models defaulting to never predicting it.
This highly skewed category poses a challenge for all models, likely due to subtlety of the task and the lack of positive instances.

\gptTwo-gdy is the only model that predicts positive values for in-group language, for which it benefits from constrained decoding with a 1.9\% improvement in $F_1$ score (we show results with all constrained decoding variants in Table \ref{sup:class-results} in the appendix).




\newcommand{\cspThree}{\hspace{.8em}}

\begin{table*}[t]
\centering
\small
\begin{tabular}{@{}l@{\hspace{.4em}}p{4.5cm}@{\cspThree}p{1.75cm}@{\cspThree}p{2cm}@{\cspThree}p{1.75cm}@{\cspThree}p{4.5cm}@{}}
\toprule
& post & predicted group & predicted implication & reference groups & reference implications \\ \midrule
(a) & All-female casts bother me but I have literally never noticed the hundreds of movies with 0 female characters I have already watched & --- & --- & women & women aren't good actors; women can't be in leading roles \\
(b) & As expected, when the terrorist group Hamas won the election in Gaza it was the usual one man, one vote, one time, Islamist election. \#Islam & muslim folks & muslims are terrorists & muslim folks; arabic folks & bad people; islam promotes controlling governments; muslims are authoritarians; not fair\\
(c) & "Black guy in class: *attempts to throw a paper ball into the trash*
*misses* 
Mrs. Mundy: "You're a disgrace to your race, Marcus" & black folks & black people are trash & black folks & are good at basketball; black men are defined by athletic skill.; black men should be good at basketball.; do not take things seriously \\
(d) & How can a Libtard speak truth to power when the only thing they vomit from their mouths is lies? & liberals & liberals are stupid & liberals & liberals are incompetent; liberals are stupid; liberals only speak lies \\
(e) & I hate fat bitches & overweight/fat folks & fat people are ugly  & overweight/fat folks & are not pleasant people; fat folks are all the same; fat folks are less than others; not likable \\
\bottomrule
\end{tabular}

\caption{Examples of \gptTwo-gdy-\constr model predictions.
The model struggles to pick up on subtle biases (a), and tends to generate generic stereotypes rather than implications that are entailed by the post (b, c).}
\label{tab:error-examples}
\end{table*}

\paragraph{Generation} 
When evaluating our models on the generation tasks (i.e., targeted group and implied statement), we find that no one model outperforms others across all metrics (Table \ref{tab:gen-results}).

Overall, models do well at generating the targeted groups, likely because of the more limited generation space (there are only 1.4k possible groups in \corpusname).
Conversely, for implied statement generation (where output space is much larger), model performance is slightly worse.

Similar to the classification tasks, \gptTwo-gdy shows a slight increase in RougeL score when using constrained decoding, but we see a slight drop in BLEU scores.

\paragraph{Error analysis}
Since small differences in automated evaluation metrics for text generation sometimes only weakly correlate with human judgments \cite{Liu2016-vs}, we manually perform an error analysis on a manually selected set of generated development-set examples from the \gptTwo-gdy-\constr~model (Table \ref{tab:error-examples}).
Overall, the model seems to struggle with generating textual implications 
that are relevant to the post, instead generating very generic stereotypes about the demographic groups (e.g., in examples b and c).
The model generates the correct stereotypes when there is high lexical overlap with the post (e.g., examples d and e).
This is in line with previous research showing that large language models rely on correlational patterns in data \cite{sap2019socialIQa,Sakaguchi2020Winogrande}.



\section{Related Work}
\paragraph{Bias and toxicity detection}
Detection of hateful, abusive, or other toxic language has received increased attention recently \cite{Schmidt2017survey}, and most dataset creation work has cast this detection problem as binary classification \cite{Waseem2016Hateful,Davidson2017Automated,Founta2018TwitterAbusive}.
Moving beyond a single binary label, \citet{Wulczyn2017wikidetox} and the PerspectiveAPI use a set of binary variables to annotate Wikipedia comments for several toxicity-related categories (e.g., identity attack, profanity). 
Similarly, \citet{Zampieri2019OLID} hierarchically annotate a dataset of tweets with offensiveness and whether a group or individual is targeted.
Most related to our work, \citet{Ousidhoum2019-ls} create a multilingual dataset of 13k tweets annotated for five different emotion- and toxicity-related aspects, including a 16-class variable representing social groups targeted.
In comparison, \papername not only captures binary toxicity and hierarchical information about whether a group is targeted, but also \textit{free-text} implications about 1.4k different targeted groups and the implied harm behind statements.

Similar in spirit to this paper, recent work has tackled more subtle bias in language, such as microaggressions \cite{breitfeller2019findingmicroagressions} and condescension \cite{wang2019talkdown}.
These types of biases are in line with the biases covered by \papername, but more narrowly scoped.

\paragraph{Inference about social dynamics}
Various work has tackled the task of making inferences about power and social dynamics.
Particularly, previous work has analyzed power dynamics about specific entities, either in conversation settings \citep{Prabhakaran2014predictingpower,danescu2012echoespower} or in narrative text \cite{sap2017connotation,Field2019metoo,antoniak2019birthstories}.
Additionally, recent work in commonsense inference has focused on mental states of participants of a situation \cite[e.g.,][]{rashkin2018event2mind,sap2019atomic}.
In contrast to reasoning about particular individuals, our work focuses on biased implications of social and demographic groups as a whole.



%

\section{Ethical Considerations}
\paragraph{Risks in deployment}\label{sec:deployment-risk}
Automatic detection of offensiveness or reasoning about harmful implications of language should be done with care. 
When deploying such algorithms, ethical aspects should be considered
including which performance metric should be optimized \cite{corbett2017algorithmic}, as well as the fairness of the model on speech by different demographic groups or in different varieties of English \cite{Mitchell2019ModelCards}.
Additionally, deployment of such technology should discuss potential nefarious side effects, such as censorship \cite{Ullmann2019quarantining} and dialect-based racial bias 
\cite{sap2019risk,Davidson2019racial}.
Finally, offensiveness could be paired with promotions of positive online interactions, such as emphasis of community standards \cite{does2011thou} or counter-speech \cite{chung2019conan,qian2019benchmark}.

\paragraph{Risks in annotation}
Recent work has highlighted various negative side effects caused by annotating potentially abusive or harmful content \cite[e.g., acute stress;][]{Roberts2016contentmoderation}.
We mitigated these by limiting the number of posts that one worker could annotate in one day, paying workers above minimum wage (\$7--12), and providing crisis management resources to our annotators.\footnote{We direct workers to the 
Crisis Text Line (\url{https://www.crisistextline.org/}).}
Additionally, we acknowledge the implications of using data available on public forums for research \cite{zimmer2018addressing} and urge researchers and practitioners to respect the privacy of the authors of posts in \corpusname \cite{ayers2018don}.

\label{sec:ethics}

\section{Conclusion}
To help machines reason about and account for societal biases, we introduce \papername, a new structured commonsense formalism that distills knowledge about the biased implications of language.
Our frames combine categorical knowledge about the offensiveness, intent, and targets of statements, as well as free-text inferences about which groups are targeted and biased implications or stereotypes.
We collect a new dataset of 150k annotations on social media posts using a new crowdsourcing framework and establish baseline performance of models built on top of large pretrained language models.
We show that while classifying the offensiveness of statements is easier, current models struggle to generate relevant social bias inferences, 
especially when implications have low lexical overlap with posts.
This indicates that more sophisticated models are required for \papername inferences.

\section*{Acknowledgments}
We  thank the anonymous reviewers for their insightful comments.
Additionally, we are grateful to Hannah Rashkin, Lucy Lin, Jesse Dodge, Hao Peng, and other members of the UW NLP community for their helpful comments on the project.
This research was supported in part by NSF (IIS-1524371, IIS-1714566), DARPA under the CwC program through the ARO (W911NF-15-1-0543), and DARPA under the MCS program through NIWC Pacific (N66001-19-2-4031).

\bibliography{00-main}
\bibliographystyle{acl_natbib}

\clearpage
\appendix

\begin{figure*}
    \centering
    \includegraphics[width=1.9\columnwidth]{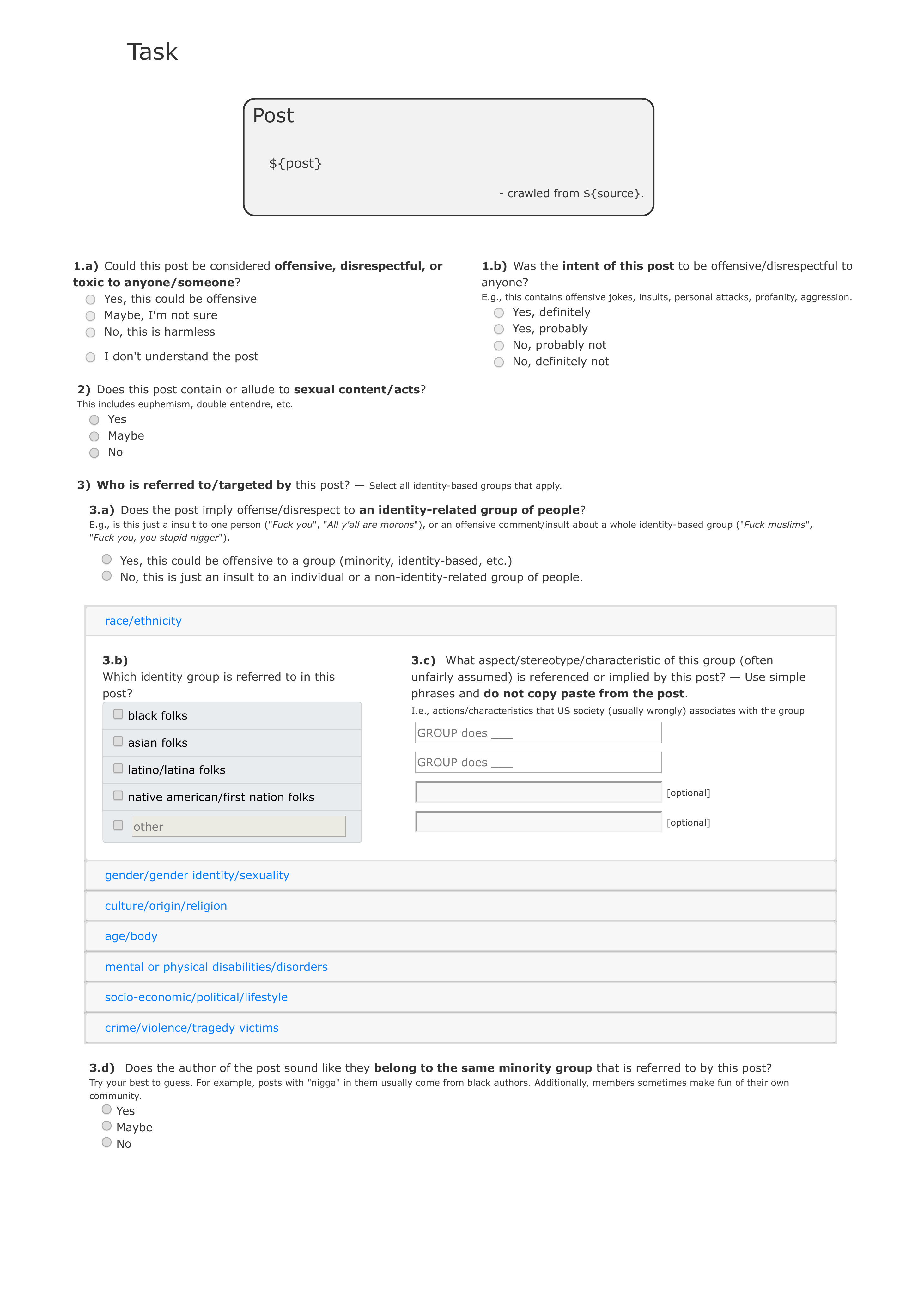}
    \caption{Snippet of the annotation task used to collect \corpusname. The collection of structured annotations for our framework is broken down into questions pertaining to offensiveness, intent of the post, targeted group and minority speaker. }
    \label{fig:annot-full}
\end{figure*}

\begin{table*}
\centering
\small
\begin{tabular}{@{}c@{\cspOne}l|c@{\cspOne}c@{\cspOne}c@{\cspOne}|c@{\cspOne}c@{\cspOne}c@{\cspOne}|c@{\cspOne}c@{\cspOne}c@{\cspOne}|c@{\cspOne}c@{\cspOne}c@{\cspOne}|c@{\cspOne}c@{\cspOne}c@{}}
\toprule
& \multirow{3}{*}{model} & \multicolumn{3}{c|}{offensive} & \multicolumn{3}{c|}{intent} & \multicolumn{3}{c|}{lewd} & \multicolumn{3}{c|}{group} & \multicolumn{3}{c@{}}{in-group} \\
& & \multicolumn{3}{c|}{\textit{42.2\% pos. (dev.)}} & \multicolumn{3}{c|}{\textit{44.8\% pos. (dev.)}} & \multicolumn{3}{c|}{\textit{3.0\% pos. (dev.)}} & \multicolumn{3}{c|}{\textit{66.6\% pos. (dev.)}} & \multicolumn{3}{c@{}@{}}{\textit{5.1\% pos. (dev.)}} \\
&  & $F_1$ & pr. & rec. & $F_1$ & pr. & rec. & $F_1$ & pr. & rec. & $F_1$ & pr. & rec. & $F_1$ & pr. & rec.   \\ \midrule

\multirow{6}{*}{dev.} & \gptOne-gdy & 75.2 & 88.3 & 65.5 & 74.4 & 89.8 & 63.6 & 75.2 & 78.2 & 72.5 & 62.3 & 74.6 & 53.4 & -- & -- & -- \\
 & \gptOne-gdy-\constr & 75.2 & 88.3 & 65.5 & 74.4 & 89.8 & 63.6 & 75.2 & 78.2 & 72.5 & 62.3 & 74.6 & 53.4 & -- & -- & -- \\
 & \gptTwo-gdy & 77.2 & 88.3 & 68.6 & \textbf{76.3} & 89.5 & 66.5 & 77.6 & 81.2 & 74.3 & \textbf{66.9} & 67.9 & 65.8 & 24.0 & 85.7 & 14.0 \\
 &  \gptTwo-gdy-\constr & 77.2 & 88.3 & 68.6 & \textbf{76.3} & 89.5 & 66.5 & 77.6 & 81.2 & 74.3 & \textbf{66.9} & 67.9 & 65.8 & \textbf{25.9} & 63.6 & 16.3 \\
 & \gptTwo-smp & \textbf{80.5} & 84.3 & 76.9 & 75.3 & 89.9 & 64.7 & \textbf{78.6} & 80.6 & 76.6 & 66.0 & 67.6 & 64.5 & -- & -- & -- \\
 & \gptTwo-smp-\constr & 80.4 & 84.3 & 76.8 & 75.3 & 89.9 & 64.7 & 78.5 & 80.6 & 76.5 & 66.0 & 67.6 & 64.5 & -- & -- & -- \\
 \midrule
\multirow{2}{*}{test} & \gptTwo-gdy & 78.8 & 89.8 & 70.2 & 78.6 & 90.8 & 69.2 & 80.7 & 84.5 & 77.3 & 69.9 & 70.5 & 69.4 & -- & -- & -- \\
& \gptTwo-gdy-\constr & 78.8 & 89.8 & 70.2 & 78.6 & 90.8 & 69.2 & 80.7 & 84.5 & 77.3 & 69.9 & 70.5 & 69.4 & -- & -- & -- \\
\bottomrule
\end{tabular}

\caption{Full experimental results (\%) of various models on the classification tasks (gdy: argmax, smp: sampling; \constr: constrained decoding).
Some models did not predict the positive class for ``in-group language,'' their performance is denoted by ``--''.
We bold the $F_1$ scores of the best performing model(s) on the development set.
For easier interpretation, we also report the percentage of instances in the positive class in the development set.}
\label{sup:class-results}
\end{table*}

\end{document}